# Omnidirectional Tractable Three Module Robot

Kartik Suryavanshi, Rama Vadapalli, Ruchitha Vucha, Abhishek Sarkar and K Madhava Krishna

*Abstract*— This paper introduces the Omnidirectional Tractable Three Module Robot for traversing inside complex pipe networks. The robot consists of three omnidirectional modules fixed 120° apart circumferentially which can rotate about their own axis allowing holonomic motion of the robot. The holonomic motion enables the robot to overcome motion singularity when negotiating T-junctions and further allows the robot to arrive in a preferred orientation while taking turns inside a pipe. We have developed a closed-form kinematic model for the robot in the paper and propose the 'Motion Singularity Region' that the robot needs to avoid while negotiating T-junction. The design and motion capabilities of the robot are demonstrated both by conducting simulations in MSC ADAMS on a simplified lumped-model of the robot and with experiments on its physical embodiment.

## I. Introduction

There have been many research efforts towards the development of a robot for in pipe locomotion in complex pipe networks [1-24]. Some of the early works involve the development of passively driven Pipe Inspection Gauges (PIGs) [1-2]. Later, many configurations were explored including the 'Articulated Snake' type which includes the THES [3], the PIPETRON series [4], PIRATE [5], the AIRO series [20-21] where the driving modules were axially separated by some distance. Screw design was explored in the development of Screw drive robot [6] and Steerable Screw robot [24] for a simpler and more compact design.

The configurations discussed above have driving modules axially separated or consist of distinct modules for driving and pathway selection. The robot configuration with three driving modules fixed 120° apart circumferentially have been extensively researched. These robots are dynamically stable, have superior mobility and a closed-form kinematic relation can be easily derived for such a configuration. MRINSPECT series has been the most successful robot series with this configuration [8-12], the latest of which has a 3D mechanical speed differential, brake, and clutch mechanism for ease of retrieval. Other research works involve modifications to this configuration for additional capabilities [13-15].

Despite these benefits, robots with three driving modules fixed circumferentially sometimes fail to negotiate the T-junction. This inability was discussed in the development of Two-Module Collaborative indoor pipeline Robot by Young, et al [16-17]. This was phrased as 'Motion Singularity' in their work. Jong, et al also discussed motion singularity in the development of FAMPER [18] and proposed a two-caterpillar mechanism to negotiate T-junction.

We propose a robot with three Omni-directional driving modules fixed on tractable arms inspired by the work of Tadakuma, et al [19]. Our robot has the novel ability to rotate in-plane about the central axis of the robot to overcome motion singularity in T-junction described later in the paper. The holonomic motion also allows the robot to come to a standard preferred orientation relative to the turning direction while negotiating elbows such that the speeds in the driving module could be modulated to some standard values according to the geometry of the turn.

This paper has been organized as follows. Section II gives a description of the robot. Section III presents the kinematic model. Section IV proposes the singularity region. Section V discusses the simulation and at the end, Section VI presents the implementation and experimentation results.

## II. Description of the Robot

### A. Robot Structure

The robot consists of three omnidirectional modules fixed 120° apart circumferentially on triangular-shaped center chassis (Fig.2(a)). Each module slide on four shafts and are pushed radially outward by linear springs secured outside the shafts. The slots in module arms are slightly bigger than the shaft which provides additional tolerance for asymmetrical compression for turning in elbows and T-junctions (Fig.2 (b)). The shaft caps attached at the ends of shafts restrict the module arm motion within the shafts. The screw slots in the center chassis allow for manual screwing of open-ended springs in the slot to adjust the compressible height of springs. The length and diameter of the robot are carefully calculated based on work in [8], for negotiating elbows and T-junctions of standard pipe with 160 mm inside diameter.

### B. Module

Each module consists of two motors one for driving the

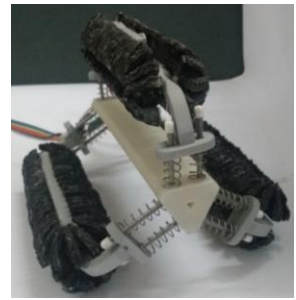

Fig.1: Omnidirectional Tractable Three Module Robot.

All the authors are with Robotics Research Center, International Institute of Information Technology, Hyderabad, India. 500032. (e-mail: suryavanshikartik@gmail.com)

two crawlers and other for rotation of modules about their central axis. The module components are described in the following subsections.

*1) Crawler Motor Assembly*

A single motor is used to drive two crawler chains (Fig.3). The motor is fixed along the module central axis and motor power is transmitted by a set of bevel gear to the sprocket assembly. The sprockets are locked to the rotating shaft which runs the two front sprockets simultaneously. Passive sprockets (Fig.4) are used at the other end of the module to complete the sprocket chain loops.

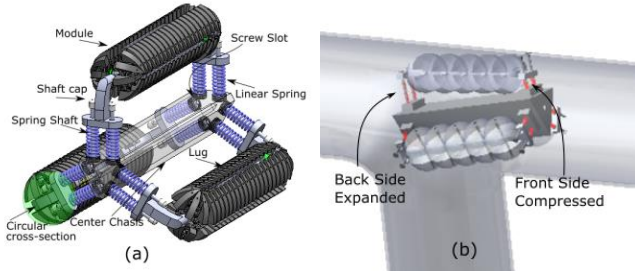

Fig.2: (a) Omnidirectional Three Module Tractable Robot (b) Asymmetrical compression during turning.

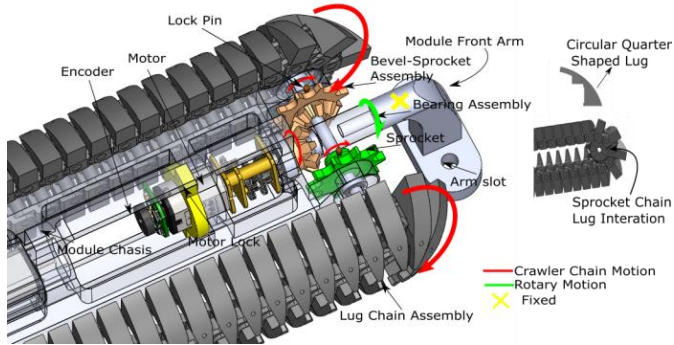

Fig.3: Exploded view of the front of the module.

*2) Rotary Motor*

The rotary motor is fixed inside the module and its shaft is rigidly fixed with the rear arm of the module (Fig.4). When the motors are run in the clockwise direction the motor body rotates in the anti-clockwise direction which in turn rotates the module in the anti-clockwise direction. Fixing the motor in this manner leaves space for the crawler chains to be brought closer such that the module size can be made compact.

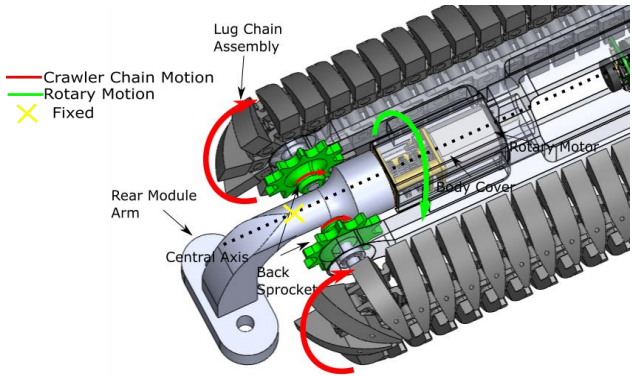

Fig.4: Exploded view of the rear of the module.

*3) Crawler Chain Assembly*

Each module has two crawler chain assemblies. The chains consist of circular quarter shaped studs called lugs (Fig.3). The design of lugs is such that the module gets a circular cross-section. Circular cross-section enables more contact at the pipe surface and also allows holonomic motion of the robot.

### III. KINEMATIC MODEL

TABLE 1
DEFINITION OF THE VARIABLES IN THE TEXT

| | |
|---|---|
| $x_g y_g z_g$ | Global frame |
| $x_r y_r z_r$ | Robot frame (rotating, non-inertial) |
| $x_{ri} y_{ri} z_{ri}$ | Robot frame (non-rotating, inertial) |
| $\dot{\theta}_1, \dot{\theta}_2, \dot{\theta}_3$ | Angular velocity of individual motors |
| $\dot{\theta}_4$ | The holonomic motion of the robot along the z-direction (Angular velocity of $x_r y_r z_r$) |
| $\theta_5$ | The angle relative to the turning direction |
| $\omega_{z1}, \omega_{z2}, \omega_{z3}$ | Module angular velocity about their central axis |
| $V_1, V_2, V_3$ | The translational velocity of modules |
| $V_{cz}$ | The translational velocity of the robot in the z-direction |
| $r$ | Lug radius |
| $f$ | Any general vector function in the rotating non-inertial frame. |
| $l_1, l_2, l_3$ | The distance of module center from the center of the robot |
| $a$ | The perpendicular distance between the center of the robot and the line joining the other two modules |
| $\omega_x, \omega_y, \omega_z$ | Angular velocity of the robot in the x, y, z-direction |
| $\omega_a$ | Robot input velocity vector |
| $V_a$ | Robot output velocity vector |
| $[J_a^u]$ | Jacobian relating input to the output velocity vector |
| $D$ | The pipe diameter |

The global reference frame $x_g y_g z_g$ and the local reference frame $x_r y_r z_r$ are shown in Fig.5. The local frame is fixed at the center of the head of the robot such that the x-axis always points towards the module with velocity $V_1$. $\hat{\imath}$, $\hat{\jmath}$ and $\hat{k}$ are the unit vectors in the local frame. The modules can translate along the global z-axis and rotate about their own axis.

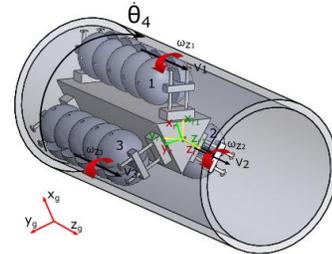

Fig.5: Frame $x_r y_r z_r$ and $x_{ri} y_{ri} z_{ri}$ are instantaneously co-aligned such that the $f(t)$ is the same in both frames and have been shown with a slight offset just for emphasis. The picture also shows module velocities in the pipe climbing robot.

Fig.5 shows the translational ($V_1$, $V_2$, $V_3$) and rotational velocities ($\omega_{z1}, \omega_{z2}, \omega_{z3}$) for the omnidirectional modules. Assuming that the lugs do not slip and maintain line contact, the linear velocities are given by

$$V_1 = r\dot{\theta}_1 \tag{1}$$
$$V_2 = r\dot{\theta}_2 \tag{2}$$
$$V_3 = r\dot{\theta}_3 \tag{3}$$

The rotation of the modules about their own axis causes the robot to rotate about the local z-axis in the opposite direction. $\dot{\theta}_4$ is the angular velocity of robot rotation between the local reference frame $x_r y_r z_r$ and global reference frame $x_g y_g z_g$ about the z-axis. Therefore, the translation vector of each module experiences a Coriolis effect due to the rotation of the robot. To find the effect of rotation on vectors, we take another inertial frame $x_{ri} y_{ri} z_{ri}$ co-located with $x_r y_r z_r$. This frame translates with the robot but does not rotate with the non-inertial local frame $x_r y_r z_r$. We take in the Coriolis effect by transforming the velocities from the robot local frame to the frame $x_{ri} y_{ri} z_{ri}$. Coriolis's theorem on vectors is used to find the velocity vector in the frame $x_{ri} y_{ri} z_{ri}$

$$\left(\frac{df}{dt}\right)_{x_{ri} y_{ri} z_{ri}} = \left(\frac{df}{dt}\right)_{x_r y_r z_r} + \dot{\theta}_4 \times f(t) \tag{4}$$

For instantaneous vector $V_1$,

$$(V_1)_{x_{ri} y_{ri} z_{ri}} = \dot{\theta}_4 \hat{k} \times l_1 \hat{\imath} + V_1 \hat{k}$$
$$= \dot{\theta}_4 l_1 \hat{\jmath} + V_1 \hat{k} \tag{5}$$

Similarly, for $V_2$ and $V_3$,

$$(V_2)_{x_{ri} y_{ri} z_{ri}} = -l_2 \cos 30\, \dot{\theta}_4 \hat{\imath} - l_2 \sin 30\, \dot{\theta}_4 \hat{\jmath} + V_2 \hat{k} \tag{6}$$

$$(V_3)_{x_{ri} y_{ri} z_{ri}} = l_3 \cos 30\, \dot{\theta}_4 \hat{\imath} - l_3 \sin 30\, \dot{\theta}_4 \hat{\jmath} + V_3 \hat{k} \tag{7}$$

Taking the average of the module speeds about the origin of $x_r y_r z_r$ to get the linear velocity of robot,

$$\frac{\left((V_1)_{x_{ri} y_{ri} z_{ri}} + (V_2)_{x_{ri} y_{ri} z_{ri}} + (V_3)_{x_{ri} y_{ri} z_{ri}}\right)}{3} = \frac{(\dot{\theta}_4 l_1 \hat{\jmath} + V_1 \hat{k} - l_2 \cos 30\, \dot{\theta}_4 \hat{\imath} - l_2 \sin 30\, \dot{\theta}_4 \hat{\jmath} + V_2 \hat{k})}{3} + \frac{(l_3 \cos 30\, \dot{\theta}_4 \hat{\imath} - l_3 \sin 30\, \dot{\theta}_4 \hat{\jmath} + V_3 \hat{k})}{3} \tag{8}$$

Considering equal arm lengths $l_1 = l_2 = l_3$, simplifying the above equation, one gets,

$$\frac{\left((V_1)_{x_r y_r z_r} + (V_2)_{x_r y_r z_r} + (V_3)_{x_r y_r z_r}\right)}{3} = V_{cz} \tag{9}$$

So, the linear velocity at the center would just be the average of the sum of translation velocities of the modules. No Coriolis effect would be observed when the arm lengths are equal like in the case of holonomic motion during the vertical climb. The robot gets non-planar velocities which lead to additional stresses if the holonomic motion is given during horizontal traversal ($l_1 \neq l_2 \neq l_3$).

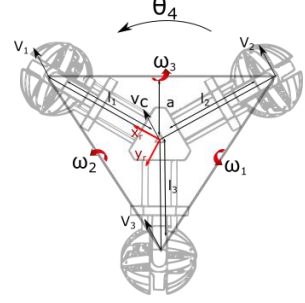

Fig.6: Velocity triangle for the robot. The translatory velocities of the modules and the robot velocity ($V_1$, $V_2$, $V_3$, $V_{cz}$) are coming out of the plane.

To calculate the angular velocity of the robot, we extrapolate the work in [16] where we also include the rotation rate of the frame $x_r y_r z_r$ ($\dot{\theta}_4$). When the other two modules are fixed and we just have $V_1$, angular velocity due to $V_1$ in the local frame $x_r y_r z_r$ is along the line joining the other two modules (Fig.6). The angular velocity due to the translation of the first module when $l_1 = l_2 = l_3 = l$ is

$$\omega_1 = -\frac{r\dot{\theta}_1}{a+l}\hat{\jmath} \tag{10}$$

Similarly, for the second and the third module,

$$\omega_2 = \frac{r\dot{\theta}_2(-\cos 30\,\hat{\imath} + \sin 30\,\hat{\jmath})}{a+l} \tag{11}$$

$$\omega_3 = \frac{r\dot{\theta}_3(\cos 30\,\hat{\imath} + \sin 30\,\hat{\jmath})}{a+l} \tag{12}$$

There also exists a component along the z-direction due to the holonomic motion of the robot which is independent of the translation velocity of the modules (Fig.6),

$$\omega_4 = \dot{\theta}_4 \hat{k} \tag{13}$$

Adding (11), (12), (13), putting $a = \frac{l}{2}$ and simplifying to get angular velocities in $x, y, z$ directions.

$$\omega_x = \left(-\frac{\sqrt{3}}{3l} r\dot{\theta}_2 + \frac{\sqrt{3}}{3l} r\dot{\theta}_3\right) \hat{\imath} \tag{14}$$

$$\omega_y = \left(-\frac{2}{3l} r\dot{\theta}_1 + \frac{1}{3l} r\dot{\theta}_2 + \frac{1}{3l} r\dot{\theta}_3\right) \hat{\jmath} \tag{15}$$

$$\omega_z = \dot{\theta}_4 \hat{k} \tag{16}$$

The relationship between input angular velocities vectors $\omega_a = (\dot{\theta}_1, \dot{\theta}_2, \dot{\theta}_3, \dot{\theta}_4)^T$ and $V_a = (\omega_x, \omega_y, \omega_z, V_{cz})^T$ the output velocity vector is defined as

$$V_a = [J_a^u]\, \omega_a \tag{17}$$

Where, $[J_a^u]$ is given as

$$[J_a^u] = \begin{bmatrix} 0 & -\sqrt{3}r/3l & \sqrt{3}r/3l & 0 \\ -2r/3l & r/3l & r/3l & 0 \\ 0 & 0 & 0 & 1 \\ r/3 & r/3 & r/3 & 0 \end{bmatrix} \quad (18)$$

The speed in the modules is set using the above equation if the geometry of the pipe is known. Assuming no slippage, the robot follows a curve if the velocities are different. The instantaneous radius of curvature of the curve is given by

$$R = \frac{|V_1| + |V_2| + |V_3|}{3\sqrt{\omega_x^2 + \omega_y^2 + \omega_z^2}} \quad (19)$$

## IV. MOTION SINGULARITY REGION

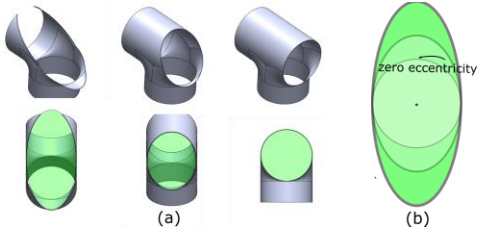

Fig.7: (a) Cross-section of the cut pipes (b) Changing eccentricity of the ellipse along the pipe.

When the robot is turning in the T-junction it encounters different elliptical pipe cross-sections of varying eccentricities as shown in Fig.7. If the orientation of the robot is such that one of the modules loses contact with the pipe surface because of the elliptical cross-section it is defined as the 'Motion singularity' [17]. In case of motion singularity, only two modules would remain in contact with the pipe surface and hence the robot would not have sufficient traction to negotiate the turn as shown in the illustrations in Fig.8(a). We find the motion singularity in the elliptical cross-section based on the intersection points of the circular geometry of the robot and the elliptical cross-section of the pipe. The region is termed as the 'Motion Singularity Region'. A sector is formed when the projection of the region is taken onto the circular cross-section of the pipe at the end of the turn (Fig.9). The detailed calculation of the motion singularity region is uploaded at https://robotics.iiit.ac.in/Archives/motion-singularity-region.pdf

For our robot, we find the angle subtended by the singularity sector in the circle is equal to 96.54°. Which gives 11.78° space for the robot to rotate in either side from the preferred orientation as shown in the Fig.9. The robot modules have to avoid the singularity region to be able to negotiate the T-junction as shown in Fig.8(b).

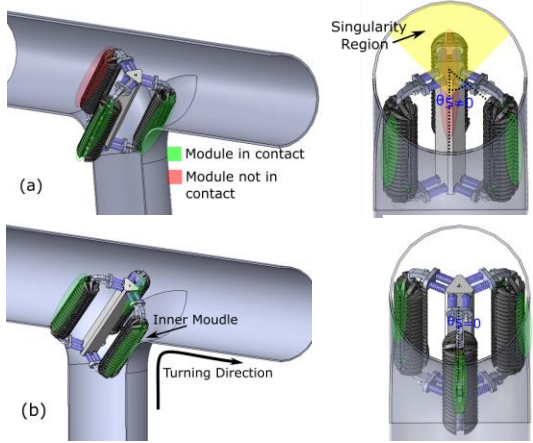

Fig.8: While negotiating the T-junction the robot can have different orientations with respect to the turning direction (a) illustration shows the worst orientation for the robot and (b) the best orientation where one module aligns with the turning direction.

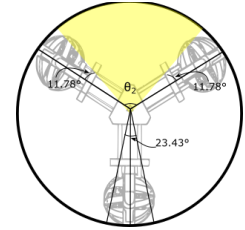

Fig.9: The yellow shaded region is the singularity region for the robot. The module has to avoid the region which leaves only 11.78° rotation angle on either side from the best orientation.

## V. MOTION SIMULATION

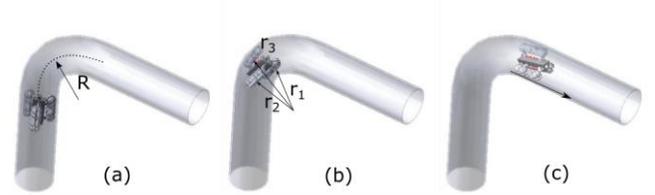

Fig.10: Robot simulation in Elbow (a) Vertical Climb (b) Modules speed modulated according to $r_1$, $r_2$, $r_3$ (c) Straight Drive.

Simulations were conducted to test the turn negotiation capability of the robot in standard elbows and T-junctions of pipe with an inner diameter of 160 mm. The analysis was conducted on a lumped model of the robot in MSC Adams. In the lumped model the complex design of the robot was simplified and the lug chain assembly was replaced by hemispherical balls to reduce the computational load in the simulation. Although the lumped model simulation could not capture the actual dynamics of the real system, it gave us instrumental design and kinematic insights.

A speed of 100 mm/s is maintained in each module to travel in straight sections. While turning in elbows (Fig.10), speeds were adjusted in the ratio of the traveling distance of the modules. The ratios are proportional to the ratios of the radius of curvature which the modules follow while negotiating the elbow $r_1: r_2: r_3$ [23] shown in the Fig.10(b).

$$\begin{aligned} V_1:V_2:V_3 &= r_1:r_2:r_3 \\ &= 1.5D - 0.5D\cos\theta_5 : 1.5D - 0.5D\cos(\theta_5 - 120°) : \\ &\quad 1.5D - 0.5D\cos(\theta_5 + 120°) \end{aligned} \quad (20)$$

For T-junctions (Fig.12), the robot is first given a holonomic (Fig.11) motion to avoid the singularity region and to orient a module in the turning direction and then driven forward. While turning in T-junction, the inner module is given speed in the backward direction and the outer modules in forward, which enables the robot to take a sharper turn. Different module speeds are initiated when the robot head reaches half of the radius of the pipe.

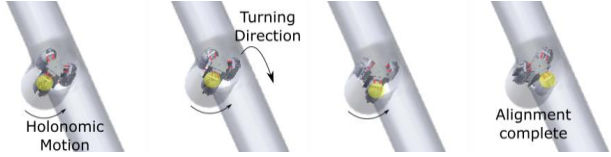

Fig.11: Snippets from the simulation of holonomic motion to align a module in the turning direction

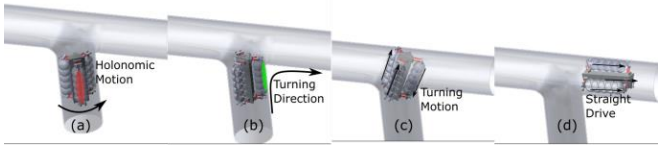

Fig.12: Simulation of the robot negotiating T-junction (a) Module is not in line with the turning direction and so is given holonomic motion (b)Module aligned with the turning direction (c) Turning motion initiated (d) Turning motion terminated.

Although the speeds are modulated according to the geometry of the turn, slippage in the inner module is observed while turning. It is difficult to estimate the exact traction, area in contact, normal force distribution and numerous other factors which makes the exact speed calculation very difficult and hence slippage cannot be eliminated completely.

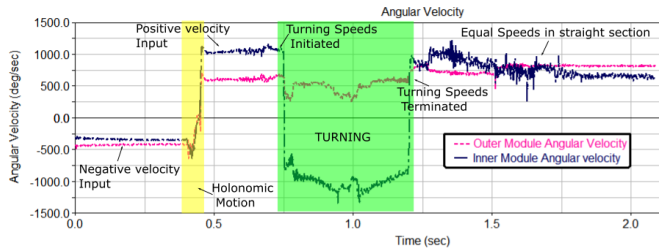

Fig.13: Angular velocity for the outer and inner module in T-junction. The yellow shaded region is where the robot is given holonomic motion and green region when turning velocities are inputted.

It was observed in the simulation that when the modules are rotated by more than 180° about their central axis the robot starts crawling in the opposite direction. This happens because the crawler chain runs in opposite directions in the upper and lower part, which explains why we have changed the input velocity in the plot after the holonomic motion in Fig.13.

## VI. IMPLEMENTATION, EXPERIMENTATION, AND RESULTS

The robot modules are designed around the smallest available motors (Pololu 1000:1 Micro Metal Gearmotor HPCB 12V with Extended Motor Shaft [25]) such that the module size remains small. The small size of the modules allows for a higher range of pipe diameter compliance. Springs with 0.5 N/mm stiffness is selected and compressible length is set by screwing in the open-ended springs in the screw slots in the center chassis. All the structural components are 3D printed with ABS plastic. Shafts are made with stainless steel to reduce the friction between moving components. The lugs are coated with latex rubber which provides sufficiently large traction to climb vertically.

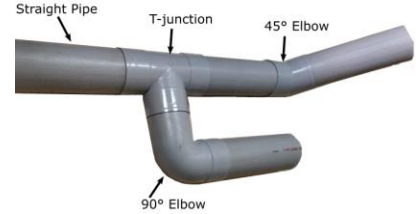

Fig.14: Picture shows different elements of a complex pipe network consisting of straight pipes, T-junction and elbows.

The robot is tested in standard pipe diameters with 160 mm inner diameter and 90 mm radius of curvature for each elbow (Fig.14). The robot is controlled at pre-defined module speeds according to the coded values. Module speeds are inputted in the code by calculating the inverse of the Jacobian derived in Section III. For negotiating an elbow, the robot is driven straight till it touches the curvature and then the turning velocities are inputted. The turning velocities are set according to the orientation of the robot relative to the turning direction. The velocities imparted are proportional to the radius of curvatures for respective modules (Eq.20). It was observed that the robot negotiates the turn more smoothly when the orientation of the robot is such that a module takes the smallest radius of curvature compared to other orientations (Fig.15). We consider this as the preferred orientation for our robot.

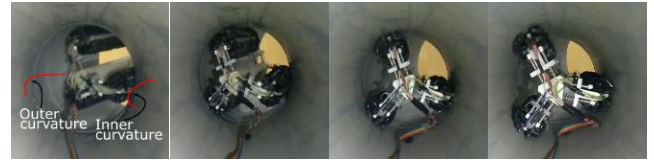

Fig.15: Robot traversing in an elbow with the preferred orientation where one module takes the innermost curvature.

The motion is smoother because while turning, the inner module is given negative speed which sometimes slips or stalls due to excessive torque as the robot is taking the turn. This exacerbates when two modules are taking a shorter radius of curvature and the slips/stalls in the modules sometimes inhibit smooth motion in the turn. Also, by bringing the robot to the preferred orientation just before reaching the turn, standard pre-calculated module speeds can be inputted according to the geometry of the turn, based on just two variables, the radius of curvature and the pipe angle. Calculating the velocities in 3D using the Jacobian for each robot orientation is not required which reduces slip in modules and reduces the computation load on the user.

For negotiating the T-junction (Fig.16), the robot module has to avoid the singularity region introduced in Section IV. The robot is given holonomic motion if a module is in the singularity region. Although the best orientation would be when the one robot module completely aligns with the turning direction, the robot is able to negotiate the turn even if the robot is not at the best orientation but all the modules are out of the singularity region. The turning speeds are initiated when the robot head reaches a quarter of the diameter of the pipe and terminated when the robot completes the turn.

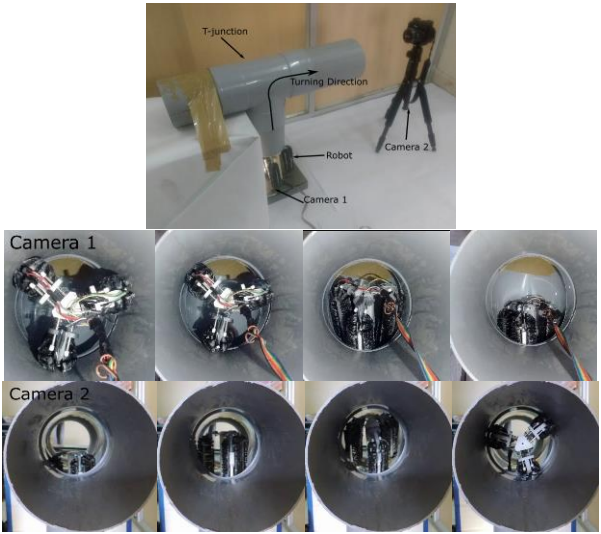

Fig.16: The experiment which involves the robot first getting to the best orientation then climbing vertically and later taking a right turn by speed modulation.

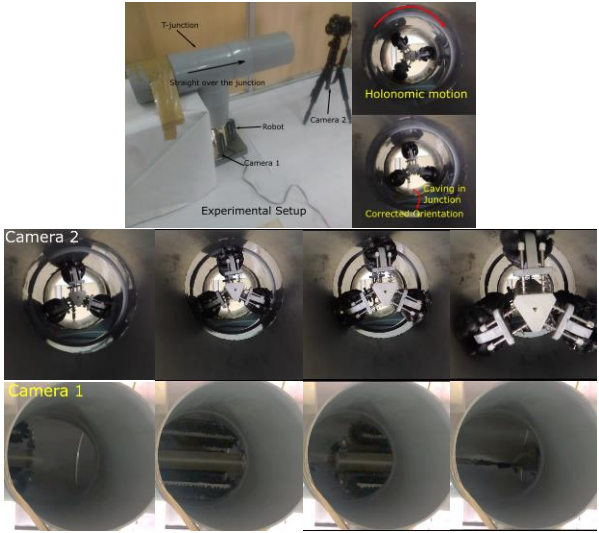

Fig.17: Experiment which involves robot correcting its orientation and then goes straight over a T-junction.

In the Fig.9, since the angle subtended by the singularity region is 96.54° out of the 120° rotation available to the module (beyond 120° the robot orientation repeats), it can be postulated that for a robot with equal dimensions but without the holonomic capability the chances of failure is 80.45% (96.54/120) or 19.55% success rate. To test this empirically, our robot was made to go through the T-junction at various angles relative to the turning direction such that one module is inside the singularity region which corresponds to situations a robot without holonomic capability woud face. It was observed that the robot sometimes falls to place to make a three-point contact with the pipe despite starting with one module inside the singularity region. However, the chances of falling to make three-point contact with the pipe decrease as the orientation of the robot gets closer to the worst orientation (or with an increase in $\theta_5$) (Fig.8(a)). These empirical estimates suggest that the success rate is higher than the postulation above. But in the case of our robot with the holonomic motion capability, the robot was able to negotiate T-junction in a clean dry plastic pipe every single time as the module can rotate away from the motion singularity region.

The robot was also tested for its capability to go straight over the T junction (Fig.17). The robot is first oriented such that the modules are on either side of the caving in T junction and then driven forward. The maneuver to orient the robot allows the robot to go more smoothly over the T junction compared to other robots.

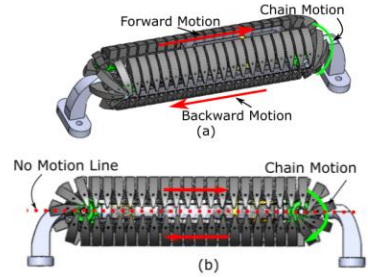

Fig.18: (a) Opposite motion above and below (b) No-Motion Line.

The opposite crawler motion is observed when the modules rotate by more than 180° consistent with the simulation (Fig.18(a)). It was also observed that when the robot module lies on the no motion line the robot just wobbles in the spot without moving forward (Fig.18(b)). This happens because the crawler chain rotates in opposite directions in the upper and the lower part of the chain. This case is avoidable as it occurs only at one angle (90°).

## VII. CONCLUSION AND FUTURE WORK

This paper introduces the Tractable Three Module Omni-crawler robot. The motion capabilities of the robot allow the robot to negotiate T-junction and tackle the problem of 'Motion Singularity'. The kinematics of the robot is formulated in the paper and the singularity region in the elliptical cross-section is identified. Simulations are conducted on a simplified lumped model which exhibit the elbow and the T-junction negotiation capability of the robot. In the end, experiments are conducted which validate the simulations. At present, we are working on making the robot autonomous which uses camera feedback for automatic orientation of the robot to the preferred orientation while taking turns in curves and T-junction based on [22] which is beyond the scope of this paper. We are also developing a 1-input 3-output differential for passive speed differential to reduce slip while turning in elbows.

## REFERENCES

[1] Roslin, Nur Shahida, et al. "A review: hybrid locomotion of in-pipe inspection robot." Procedia Engineering 41 (2012): 1456-1462.

[2] Okamoto Jr, Jun, et al. "Autonomous system for oil pipelines inspection." *Mechatronics* 9.7 (1999): 731-743.

[3] Hirose, Shigeo, et al. "Design of in-pipe inspection vehicles for/spl phi/25,/spl phi/50,/spl phi/150 pipes." Robotics and Automation,1999. Proceedings. 1999 IEEE International Conference on. Vol. 3. IEEE, 1999.

[4] P. Debenest, M. Guarnieri and S. Hirose, "PipeTron series - Robots for pipe inspection," *Proceedings of the 2014 3rd International Conference on Applied Robotics for the Power Industry*, Foz do Iguassu, 2014, pp. 1-6.

[5] Dertien, Edwin, et al. "Design of a robot for in-pipe inspection using omnidirectional wheels and active stabilization." *2014 IEEE International Conference on Robotics and Automation (ICRA)*. IEEE, 2014.

[6] Kakogawa, Atsushi, Taiki Nishimura, and Shugen Ma. "Development of a screw drive in-pipe robot for passing through bent and branch pipes." *IEEE ISR 2013*. IEEE, 2013.

[7] Kwon, Young-Sik, et al. "A flat pipeline inspection robot with two wheel chains." *2011 IEEE International Conference on Robotics and Automation*. IEEE, 2011.

[8] Roh, Se-gon, and Hyouk Ryeol Choi. "Differential-drive in-pipe robot for moving inside urban gas pipelines." *IEEE transactions on robotics* 21.1 (2005): 1-17..

[9] Roh, Se-gon, et al. "Modularized in-pipe robot capable of selective navigation inside of pipelines." *2008 IEEE/RSJ International Conference on Intelligent Robots and Systems*. IEEE, 2008..

[10] Roh, Se-gon, et al. "In-pipe robot based on selective drive mechanism." *International Journal of Control, Automation and Systems* 7.1 (2009): 105-112..

[11] Kim, Ho Moon, et al. "An in-pipe robot with multi-axial differential gear mechanism." *2013 IEEE/RSJ international conference on intelligent robots and systems*. IEEE, 2013.

[12] Yang, Seung Ung, et al. "Novel robot mechanism capable of 3D differential driving inside pipelines." *2014 IEEE/RSJ International Conference on Intelligent Robots and Systems*. IEEE, 2014.

[13] Kwon, Young-Sik, et al. "A pipeline inspection robot with a linkage type mechanical clutch." *2010 IEEE/RSJ International Conference on Intelligent Robots and Systems*. IEEE, 2010..

[14] Park, Jungwan, et al. "Normal-force control for an in-pipe robot according to the inclination of pipelines." *IEEE transactions on Industrial Electronics* 58.12 (2010): 5304-5310.

[15] Kakogawa, Atsushi, and Shugen Ma. "Design of an underactuated parallelogram crawler module for an in-pipe robot." *2013 IEEE International Conference on Robotics and Biomimetics (ROBIO)*. IEEE, 2013.

[16] Kwon, Young-Sik, and Byung-Ju Yi. "The kinematic modeling and optimal paramerization of an omni-directional pipeline robot." *2009 IEEE International Conference on Robotics and Automation*. IEEE, 2009.

[17] Kwon, Young-Sik, and Byung-Ju Yi. "Design and motion planning of a two-module collaborative indoor pipeline inspection robot." *IEEE Transactions on Robotics* 28.3 (2012): 681-696.

[18] Kim, Jong-Hoon, Gokarna Sharma, and S. Sitharama Iyengar. "FAMPER: A fully autonomous mobile robot for pipeline exploration." *2010 IEEE International Conference on Industrial Technology*. IEEE, 2010.

[19] Tadakuma, Kenjiro, et al. "Basic running test of the cylindrical tracked vehicle with sideways mobility." *2009 IEEE/RSJ International Conference on Intelligent Robots and Systems*. IEEE, 2009..

[20] Kakogawa, Atsushi, and Shugen Ma. "Design of a multilink-articulated wheeled inspection robot for winding pipelines: AIRo-II." *2016 IEEE/RSJ International Conference on Intelligent Robots and Systems (IROS)*. IEEE, 2016.

[21] A. Kakogawa and S. Ma, Design of a Multilink-articulated Wheeled Pipeline Inspection Robot using Only Passive Elastic Joints, Advanced Robotics, Vol. 32, Iss. 1, pp. 37-50, 2018.

[22] Kakogawa, Atsushi, Yuki Komurasaki, and Shugen Ma. "Anisotropic shadow-based operation assistant for a pipeline-inspection robot using a single illuminator and camera." *2017 IEEE/RSJ International Conference on Intelligent Robots and Systems (IROS)*. IEEE, 2017.

[23] Roh, Se-gon, and Hyoukryeol Choi. "Strategy for navigation inside pipelines with differential-drive inpipe robot." *Proceedings 2002 IEEE International Conference on Robotics and Automation (Cat. No. 02CH37292)*. Vol. 3. IEEE, 2002.

[24] Li, Te, et al. "Design and locomotion control strategy for a steerable in-pipe robot." *2015 IEEE International Conference on Mechatronics and Automation (ICMA)*. IEEE, 2015.

[25] "Pololu - 1000:1 Micro Metal Gearmotor HPCB 12V with Extended Motor Shaft," Pololu Robotics & Electronics. [Online]. Available: https://www.pololu.com/product/3057. [Accessed: 15-Sep-2019].